\documentclass[10pt]{article}

\usepackage{arxiv}
\usepackage[utf8]{inputenc}
\usepackage[T1]{fontenc}
\usepackage[numbers,sort&compress]{natbib}
\usepackage{amsmath,amssymb}
\usepackage{graphicx}
\usepackage{booktabs}
\usepackage{subfig}
\usepackage{algorithm}
\usepackage{algpseudocode}
\usepackage{microtype}
\usepackage[hidelinks]{hyperref}
\usepackage{url}

\allowdisplaybreaks

\newcommand{\myargmin}[1]{\underset{#1}{\arg\!\min}\,}
\newcommand{\eq}[1]{(\ref{#1})}
\newcommand{\meanstd}[2]{#1\,{\scriptstyle \pm\,#2}}
\newcommand{\accci}[3]{$#1\,{\scriptstyle [#2,#3]}$}

\renewcommand{\headeright}{arXiv preprint}
\renewcommand{\undertitle}{Preprint}
\renewcommand{\shorttitle}{Multi-Depth Temporal Fusion for Locally Trained SNNs}

\title{Multi-Depth Temporal Fusion for Feedforward,\\
Locally Trained Spiking Neural Networks}

\author{\normalfont
\textbf{Aidin Attar}\textsuperscript{*}\\
\footnotesize Department of Information\\
\footnotesize Engineering, University of Padua\\
\footnotesize Padua, Italy
\and
\textbf{Eleonora Cicciarella}\\
\footnotesize Department of Information\\
\footnotesize Engineering, University of Padua\\
\footnotesize Padua, Italy
\and
\textbf{Michele Rossi}\\
\footnotesize Department of Information\\
\footnotesize Engineering, University of Padua\\
\footnotesize Department of Mathematics\\
\footnotesize ``Tullio Levi-Civita'', University of Padua\\
\footnotesize Padua, Italy
}

\date{}

\begin{document}
\maketitle

\begin{center}
\small \textsuperscript{*}Corresponding author: \texttt{aidin.attar@phd.unipd.it}
\end{center}
\vspace{0.35em}

\begin{abstract}
We propose a new spiking neural network (SNN) design to process static images and event streams using time-to-first-spike (TTFS) latencies. Our key research question is which architectural choices best accommodate local and online learning in multi-layer convolutional SNNs. This question is addressed via an original framework combining residual-like connections with {\it multi-depth feature aggregation} and {\it consensus}. The full SNN pipeline features an early-vision front end, to convert raw visual data into sparse spike latencies, a four-layer convolutional backbone trained layerwise with unsupervised spike-timing-dependent plasticity (STDP), a deterministic Multi-Depth Temporal Fusion (MDTF) and a final classifier trained with reward-modulated spike-timing-dependent plasticity (R-STDP). Rather than replacing early features in deeper layers, the proposed MDTF preserves early temporal evidence, adding sparse residual events from intermediate layers, and incorporating deeper features only when they agree in time with earlier representations. The resulting architecture is experimentally validated across MNIST, Fashion-MNIST, CIFAR-10, and N-MNIST, delivering strong classification performance under a fully local learning regime. Selective multi-depth fusion significantly outperforms traditional STDP/R-STDP baselines on higher-variability visual tasks (achieving $+18.2$ pp on Fashion-MNIST and $+29.2$ pp on CIFAR-10). Furthermore, activity-budget analyses show that the network retains high accuracy even when removing a large fraction of late or weak spike events, confirming its high data efficiency and reduced event-processing requirements. The codebase is publicly available at \url{https://github.com/aidinattar/multi-depth-temporal-fusion-snn}.
\end{abstract}


\section{Introduction}
\label{sec:introduction}

Artificial intelligence systems have become increasingly effective, but often at the cost of larger (deeper) models, higher training complexity, and growing energy demands at both training and inference time.
For this reason, there is sustained interest in neural models that can retain useful representational power while operating under tighter efficiency constraints. SNNs are natural candidates in this sense because they process information through {\it discrete events} and exploit time as an explicit computational dimension. Moreover, their sparse and event-driven nature makes them ideal for deployment on neuromorphic hardware ~\cite{roy_towards_2019,nunes_spiking_2022,davies_loihi_2018,richter_dynap-se2_2024}.

Despite these advantages, training SNNs remains nontrivial. In contrast to conventional artificial neural networks (ANNs), spiking neurons combine internal state dynamics with discontinuous spike emission, making the learning problem {\it inherently time-dependent} and overall more challenging to solve. 
A large body of work has recently addressed this through surrogate-gradient methods, which make end-to-end training possible by replacing the non-differentiable spike function with a differentiable approximation during optimization~\cite{lee_training_2018,neftci_surrogate_2019}. Another major line of work relies on direct ANN-to-SNN conversion, where a high-performing ANN model is first trained and then transferred into the spiking domain~\cite{ding_optimal_2021,deng_optimal_2021,bu_optimal_2023,hao_reducing_2023}. These approaches have produced good results. However, they leave open a complementary and yet relevant question: how much can be achieved in deep SNNs when learning remains {\it local}, online, and aligned with the constraints that originally motivated spike-based computation? With ``local'' learning rules, we mean learning dynamics where synaptic updates solely depend on pre- and post-synaptic neural activity, possibly
combined with a modulatory signal, without entailing the propagation of global error signals (e.g., backpropagation). This type of adaptation is here referred to as {\it local plasticity} and is the main focus of the present work.

Local plasticity mirrors the mechanisms that biological systems use to learn and is also attractive as a means to design online training algorithms for neuromorphic hardware, where propagating global error signals can be costly. Among local learning rules, STDP is one of the most established. It captures the dependence of synaptic change on the relative timing of pre- and post-synaptic spikes and has long been regarded as a plausible mechanism for unsupervised feature representation~\cite{bi_synaptic_1998}. However, this same locality also defines its main limitation: standard STDP strengthens responses to recurring input patterns, but provides no direct mechanism for favouring features that are useful for a specific final task.

In this respect, reward-modulated STDP (R-STDP) introduces an important extension: by coupling local synaptic eligibility with a delayed scalar {\it reinforcement signal}, it allows weight updates to remain local while still depending on the task outcome~\cite{fremaux_neuromodulated_2016}. R-STDP provides a biologically inspired form of three-factor learning that stands between purely unsupervised plasticity and fully supervised optimization. Previous work has shown that it can be effective as it promotes the learning of more discriminative features than pure STDP. Moreover, it enables classification to be performed directly in the spiking domain, eliminating the need for an external classifier~\cite{mozafari_first-spike-based_2018,mozafari_bio-inspired_2019}. These results provide evidence that local reward-based plasticity can support effective task-driven learning in practical architectures, suggesting its applicability beyond simplified experimental settings.

Most prior work on R-STDP was obtained on relatively compact architectures. However, as model depth increases, its effectiveness becomes considerably less clear. In standard deep learning (DL), the benefits of increased depth are largely enabled by architectural strategies that improve optimization and representation learning, such as residual connections and multi-scale feature aggregation~\cite{szegedy_going_2015,he_deep_2016}. Related ideas have also been shown to play an important role in modern deep~SNNs, especially in gradient-based optimization where residual connections can significantly improve the final model accuracy~\cite{fang_deep_2021}. However, it remains unclear whether residual paths and multi-branch feature aggregation retain the same value when the network is trained only through local STDP/R-STDP updates. In a deep spiking model trained with STDP or R-STDP, greater architectural complexity can improve representation learning but also exacerbates spike sparsity, unstable dynamics, and makes temporal credit assignment (TCA) more difficult. Currently, the balance between these effects is still not well understood.

In this work, we shed new light on the role of online learning rules in multi-layer convolutional SNNs, where features encode the time to first spike (TTFS) and training is achieved via local STDP and R-STDP. Our architectural design follows the objective of preserving informative spike-time evidence as it propagates through increasingly deep processing stages. Thanks to an original Multi-Depth Temporal Fusion (MDTF), such evidence is progressively refined by architecturally combining residual-inspired connections, multi-scale feature aggregation and consensus. Since learning for the internal network layers is based on STDP, without using a global error signal, any information that is suppressed by an intermediate stage cannot be directly recovered by later plasticity. This makes the representation reaching the final readout layer particularly important. Our SNN architecture is trained and tested on visual recognition tasks, considering both static images and event-based visual inputs, where increasing input variability makes the preservation and selective refinement of early temporal evidence progressively more important.

In the proposed SNN design, network topology and learning dynamics {\it are tightly coupled}: architectural choices (namely, our MDTF) determine not only which representations reach the final classifier, but also which spike-time events remain available to subsequent synaptic adaptation. We exploit this coupling explicitly, treating depth as a problem of temporal evidence routing under local learning rather than simply as the addition of further processing stages.

The proposed model is designed around this interaction between information flow and local plasticity, ensuring that useful spike-time evidence remains available as processing depth increases. The input layer, referred to as the ``front end'', is based on a functional model of early vision. It performs deterministic transformations to move visual streaming data into feature vectors encoded in terms of spike latency, and consists of four stages: \emph{local decorrelation} to reduce redundancy, \emph{polarity processing} to separate positive and negative contrasts, \emph{gain control} to stabilize response ranges, and \emph{latency coding} to convert the resulting information into sparse first-spike times ~\cite{atick_what_1992,pitkow_decorrelation_2012,ichinose_and_2022,carandini_normalization_2011,rullen_rate_2001}. 
The following convolutional layers constitute the ``SNN backbone'', trained using STDP. The deterministic MDTF adapts residual and inception-like ideas: early spike-time evidence is preserved, intermediate representations contribute through sparse residual events, and deeper evidence is retained only when it is temporally consistent with the early and intermediate representations. The final decision stage consists of a single fully connected spiking R-STDP readout with multiple prototypes per class that operates directly on the event representation produced by the backbone. The experimental evaluation then separates the roles of the visual front end, feature propagation, deeper branching, and reward-modulated decision learning. The main contributions of this work are summarized as follows:
\begin{enumerate}
    \item We propose a \emph{label-free} latency encoder that combines local decorrelation, polarity separation, contrast-context gating, and response calibration to convert local visual streaming input data into calibrated {\it first-spike latency maps}; 
    \item We design the MDTF, to selectively refine the temporal features produced by the backbone. It exploits residual and agreement-based temporal routing to preserve early spike-time evidence while allowing deeper branches to add sparse corrections to the initial code;
    \item We propose a multi-prototype R-STDP readout that performs classification entirely in the spiking domain, reinforcing target-class prototypes while selectively punishing the most competitive non-target ones;
    \item The proposed pipeline is thoroughly validated via ablation studies in four representative visual datasets (MNIST, Fashion-MNIST, CIFAR-10, and N-MNIST), highlighting the role of each architectural component. Strong classification performance is achieved, proving that selective multi-depth fusion significantly outperforms traditional STDP/R-STDP baselines on higher-variability visual tasks, achieving $+18.2$ pp on Fashion-MNIST and $+29.2$ pp on CIFAR-10. Moreover, activity-budget analyses show that the network retains high accuracy even when removing a large fraction of late or weak spike events, confirming its high data efficiency and reduced event-processing requirements. These results position our solution as a practical and effective framework for defining and training multi-layer SNNs in online learning scenarios.
\end{enumerate}

The remainder of this paper is organized as follows. Section~\ref{sec:related_work} reviews the main literature on SNN learning rules and deep spiking architectures. The proposed processing pipeline is outlined in Section~\ref{sec:processing_pipeline}, while Section~\ref{sec:signals_features} describes the spike-time signals and feature representations used by the proposed model. Section~\ref{sec:learning_framework} introduces the local learning framework, and Section~\ref{sec:experiments} presents and discusses the experimental results. Finally, Section~\ref{sec:conclusions} summarizes our findings and outlines future research directions.

\section{Related Work}
\label{sec:related_work}

Learning in SNNs is usually approached either through local plasticity rules or through optimization methods directly adapted from deep learning. The former emphasizes biological plausibility by relying on local synaptic plasticity, whereas the latter prioritizes task performance and trainability through gradient-based optimization. The present work bridges these complementary directions by investigating increasingly deep convolutional architectures while utilizing local STDP/R-STDP learning mechanisms.
In computational vision models, STDP has been used to learn selective features from unlabeled input, especially when paired with temporal coding and competition ~\cite{masquelier_unsupervised_2007,diehl_unsupervised_2015}. Convolutional STDP systems have also been studied as unsupervised visual feature learners with latency-coded input, while classification is performed by a downstream classifier ~\cite{safa_event_2022, falez_unsupervised_2019, kheradpisheh_stdp-based_2018}. Related work has connected STDP to stable Hebbian learning, temporal prediction, and probabilistic inference in competitive spiking circuits~\cite{rossum_stable_2000,rao_spike-timing-dependent_2001,kappel_stdp_2014}. These studies show that STDP can organize event-based visual representations into practical vision pipelines. However, they also expose its central limitation: because synaptic updates only rely on local temporal correlations, the learned features are not necessarily the most discriminative ones for the downstream task. 
Three-factor and reward-modulated rules address this limitation by combining local eligibility with a modulatory signal that carries information about outcome, error, or reward ~\cite{izhikevich_solving_2007,legenstein_learning_2008,fremaux_neuromodulated_2016,kusmierz_learning_2017_fixed}. With this approach, reward shapes synaptic changes that are still expressed through spike timing. R-STDP is therefore well matched to task-driven spiking systems with local updates and event-based computation. 
Beyond static image recognition, local plasticity and event-driven SNNs have also been studied in settings where temporal processing and online adaptation are sought, including neuromorphic sensing, event-based datasets, reservoir computing, and robotic control~\cite{davies_loihi_2018,orchard_converting_2015,patino-saucedo_liquid_2022,tsang_radar-based_2021}. These application domains differ substantially in their input statistics, temporal structure, and task objectives, making comparisons across them difficult. We therefore restrict our experimental study to visual classification, considering both static and event-based inputs under a common recognition setting, so that the effects of temporal encoding, local plasticity, and architectural routing can be examined without conflating them with differences in task formulation.

In vision models based on first-spike timing, reward modulation makes it possible to keep the computation fully spiking while biasing plasticity toward task-relevant features. R-STDP has been shown to improve over unsupervised STDP in object categorization~\cite{mozafari_first-spike-based_2018} and was later incorporated into a convolutional architecture for digit recognition, where early layers were trained with STDP and deeper layers with reward-modulated plasticity~\cite{mozafari_bio-inspired_2019}. These works are particularly relevant because classification remains entirely within the spiking network, without requiring a separate analog classifier. However, the reward signal can modulate only the spike-time features that reach the decision layer; information suppressed by earlier locally trained stages cannot be recovered through global error propagation. As inputs and architectures become richer, preserving the structure and timing of this evidence therefore becomes increasingly important.

Under these settings, little is known about how intermediate spike-time representations are to be routed in deeper SNNs. In fact, under local plasticity, a deeper layer cannot rely on a global error signal to recover information that has already been suppressed. This makes the preservation and selective enrichment of early temporal evidence a {\it central architectural problem}. In conventional deep learning, depth becomes effective when information flow is controlled through mechanisms such as residual connections and multi-branch feature extraction~\cite{szegedy_going_2015,he_deep_2016}. Similar ideas have been successful in gradient-trained SNNs, where residual paths help stabilize the training of deeper networks~\cite{fang_deep_2021}. 
Under STDP and R-STDP, however, the role of these mechanisms is different. A skip path or a branch changes which spikes arrive, when they arrive, and which events remain available to local plasticity. This is the setting considered in the present work, where we investigate how biologically inspired processing, residual connections, inception-like branching, and reward-modulated decision learning can be combined to preserve useful temporal evidence in locally trained convolutional SNNs.

\begin{figure}[t]
    \centering
    \includegraphics[width=\textwidth]{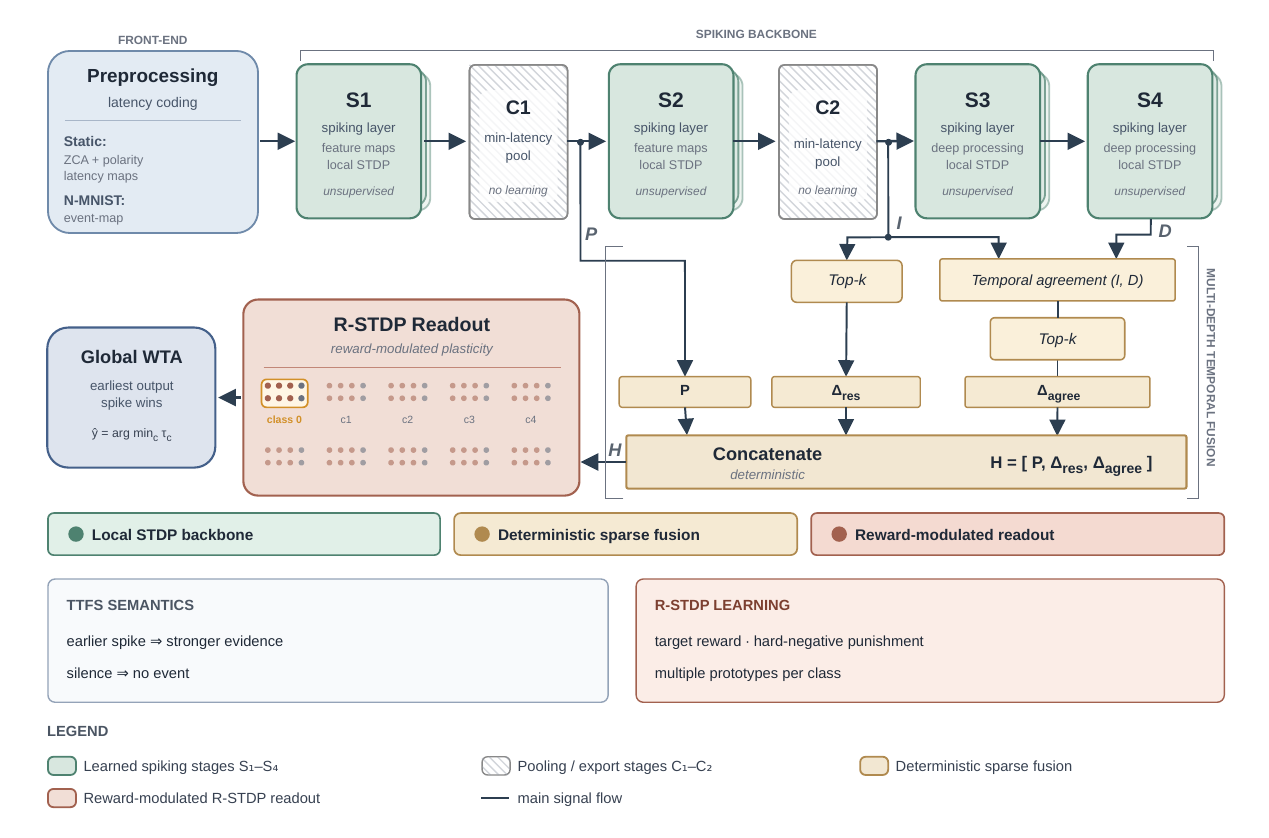}
    \caption{Schematic overview of the proposed locally trained TTFS architecture. The model uses a four-layer spiking convolutional backbone ($S_1$--$S_4$). The MDTF merges representations at different depths. It preserves the early code $P=H^{(1)}$, incorporates the sparse residual contribution $\Delta_{\mathrm{res}}$ derived from the intermediate representation $I=H^{(2)}$, and adds the agreement contribution $\Delta_{\mathrm{agree}}$ obtained from temporally consistent events in $I$ and the deep representation $D=H^{(4)}$. These components are concatenated into the final latency representation $H=[P,\Delta_{\mathrm{res}},\Delta_{\mathrm{agree}}]$, which is classified by a reward-modulated multi-prototype R-STDP readout through global winner-take-all spike timing. The earliest output spike determines the class score, silent neurons are assigned the maximum time, and membrane potential is used only for tie breaking. Labels are used only to define the reward and punishment signals at the decision stage.}
    \label{fig:general_schema}
\end{figure}

\section{Processing Pipeline Overview}
\label{sec:processing_pipeline}

The proposed model is shown in Fig.~\ref{fig:general_schema}. It is a fully SNN-based architecture organized around four core components: \textit{(i)} a label-free TTFS front end to convert images into sparse latency maps; \textit{(ii)} a stage-wise convolutional backbone to learn local temporal features via unsupervised STDP; \textit{(iii)} an original MDTF based on residual and agreement-based temporal connections preserving the stable base code, while selectively incorporating early events from deeper branches; \textit{(iv)} a multi-prototype R-STDP readout to perform the final classification. The entire pipeline is trained with local learning rules via STDP (convolutional layers S1--S4) and R-STDP (readout).

The front end defines the sensory latency code used by the stages that follow, producing a latency-based and temporally ordered representation. This module is inspired by functional early-vision processing.

The following convolutional stages (the backbone) learn local feature maps using unsupervised STDP, organizing recurring visual structure into a stable spiking code for subsequent classification. The MDTF merges deeper representations utilizing residual connections and inception-like encoding with the objective of preserving the initial code, while providing additional refinements. Under local learning, the MDTF serves a different purpose than in backpropagation-trained convolutional neural networks (CNNs): rather than facilitating gradient flow, it determines which spike-time events remain available to subsequent stages.

The final classifier is a spiking R-STDP readout trained via local spike-timing-dependent synaptic updates modulated by reward and punishment signals. Population coding is also exploited by using multiple neurons for each class.

\section{Signals and Features}
\label{sec:signals_features}

\subsection{Spike-Time Representation}

An activation in the considered model is represented as {\it the time of its first spike}. For a feature index $i$, we denote this latency by $L_i$. Earlier spikes correspond to stronger temporal evidence, while missing spikes are treated as silent features. The same convention is used for input maps, intermediate convolutional responses, residual paths, and readout inputs. This representation determines how architectural operations are defined. Concatenating feature tensors preserves multiple sources of spike-time evidence as separate channels. Selecting top-$k$ events retains only the earliest or most informative latency changes within a sample. 
Agreement between branches is also temporal: two branches are considered consistent when they produce early events at corresponding locations or feature groups whose spike times fall within a predefined temporal tolerance. These are the general principles by which the backbone manipulates latency fields directly throughout the network.

\subsection{Visual Front End and Latency Encoding}

The visual front end has been designed taking inspiration from a functional early-vision module~\cite{barlow_possible_2012}. It reshapes the input into a form that can be used by spike-timing plasticity, using operations that parallel key computational roles of early sensory pathways, namely redundancy reduction, contrast-polarity separation, local gain control, and sparse latency coding. Local decorrelation reduces input patch-level correlations in line with efficient-coding principles of early vision~\cite{barlow_possible_2012,atick_what_1992,pitkow_decorrelation_2012}. Polarity splitting treats positive and negative signals as distinct non-negative event channels, analogously to ON/OFF visual pathways~\cite{ichinose_and_2022}. Signed-context gating and polarity balancing act as deterministic contrast-normalization steps, serving the functional role of gain control without specifying a circuit-level inhibitory mechanism~\cite{heeger_normalization_1992,carandini_normalization_2011}. The final response-to-latency map then returns a sparse time-to-first-spike code, where stronger responses fire earlier and weak ones remain silent~\cite{thorpe_speed_1996,rullen_rate_2001,masquelier_unsupervised_2007}.

This initial stage establishes the spike-time representation that serves as the basis for all subsequent local plasticity mechanisms and corresponds to the \emph{Preprocessing} block in the pipeline of Fig.~\ref{fig:general_schema}. We denote the transformation carried out by this front end as
\begin{equation}
H^{(0)} = \Phi(x),
\end{equation}
where $x$ denotes the input sample and $H^{(0)}$ is the output expressed as a {\it latency tensor}, which is presented to the next convolutional spiking layer (block S1 in Fig.~\ref{fig:general_schema}). The mapping $\Phi$ is label-free. For static inputs, its data-dependent statistics are estimated once from the training split and then frozen; for event-based inputs, the transformation uses fixed sample-wise operations and no dataset-level fitted statistics. Its purpose is to transform local sensory evidence into a sparse temporal code before STDP-based learning starts.

In what follows, we detail the processing steps constituting $\Phi$. We first describe the transformations specific to static images, followed by the preprocessing adopted for event-based inputs. The resulting representations are then converted into spike times through a latency-encoding stage.

We denote the static input image tensor by
$
p\in\mathbb{R}^{C\times H\times W},
$
where $C$ is the number of image channels (one for grayscale images and three for RGB images), while $H$ and $W$ denote the spatial dimensions. 
 
\noindent \textbf{Local decorrelation.} Local patches are decorrelated through a \textit{local-decorrelation transform}, implemented as a regularized zero-phase covariance normalization whose statistics are estimated from local image patches sampled from the training set. Let $q_u\in\mathbb{R}^{C_\Pi P^2}$ be the vectorized $P\times P$ patch $\mathcal{P}$ of $p$ centered at location $u$. The empirical mean $\mu$ and covariance $\Sigma$ are computed once from patches sampled from the training data set. The local covariance-normalization matrix is
\begin{equation}
W = (\Sigma + \epsilon I)^{-1/2},
\end{equation}
where $\epsilon>0$ is a small regularization constant that improves the conditioning of $W$ and ensures a well-defined and numerically stable inverse square root. For each patch, the decorrelated patch vector is computed as
\begin{equation}
\tilde q_u = W(q_u-\mu).
\end{equation}
The vector $\tilde q_u$ is then reshaped onto the original $C_\Pi \times P \times P$ patch layout, and the value at the central spatial position of each channel $c$ is retained as the {\it signed decorrelated response}:
\begin{equation}
z_c(u),\qquad c=1,\dots,C_\Pi .
\end{equation}
Note that this operation can be equivalently implemented using a convolutional filter bank whose filters are determined by the local covariance-normalization matrix. The resulting maps have the same number of channels as the original sensory tensor, but their local second-order correlations have been reduced while preserving the spatial layout. The response $z_c(u)$ is real-valued, with magnitude measuring the strength of the local decorrelated contrast, and sign encoding the corresponding polarity. 

\noindent \textbf{Signed-context gating.} Since the following spiking layers operate on non-negative events, both signs of $z_c(u)$ must be handled explicitly prior to performing latency encoding. Before splitting the two signs into separate event maps, we apply a local {\it signed-context gate}. The gate remains weak in locally one-sided regions, while it activates when both polarities have non-negligible support, progressively biasing the response toward the locally dominant sign. Specifically, for channel $c$ and location $u$, we define the local positive and negative supports as
\begin{equation}
\begin{aligned}
S_c^+(u) &=
\sum_{v\in\mathcal{N}(u)}
\kappa(v-u)[z_c(v)]_+,\\
S_c^-(u) &=
\sum_{v\in\mathcal{N}(u)}
\kappa(v-u)[-z_c(v)]_+,
\end{aligned}
\end{equation}
where $[\cdot]_+$ denotes the positive part function, i.e., $[a]_+=\max(a,0)$ for $a\in\mathbb{R}$, $\mathcal{N}(u)$ is a neighborhood of $u$, and $\kappa(\cdot)$ is a normalized  kernel function. The degree of mixed polarity is then measured by
\begin{equation}
\rho_c(u) =
\frac{\min(S_c^+(u),S_c^-(u))}
{\max(S_c^+(u),S_c^-(u))}.
\end{equation}
When both local supports are zero, we set $\rho_c(u)=0$ by convention. Values of $\rho_c(u)$ close to zero indicate a locally one-sided response, while larger values indicate that non-negligible contributions from both signs are present. We further define the gate strength as
\begin{equation}
g_c(u) =
\alpha_{\mathrm{ctx}}\,
\operatorname{clip}\!\left(
\frac{\rho_c(u)-\rho_0}{\rho_1-\rho_0},
0,1
\right),
\end{equation}
where $\alpha_{\mathrm{ctx}}\geq 0$ controls the maximum strength, $[\rho_0,\rho_1]$ is the interval in which mixed-polarity evidence activates the gate, and $\operatorname{clip}(x,0,1)$ returns $x$ if $x\in [0,1]$, $0$ if $x<0$, and $1$ if $x>1$. 

Let $d_c(u)=\operatorname{sign}(S_c^+(u)-S_c^-(u))$ denote the locally dominant polarity around location $u$. We first define a competitive target response by preserving responses whose sign agrees with the locally dominant polarity and attenuating responses with the opposite sign by a factor $\eta\in[0,1]$:
\begin{equation}
\bar z_c(u) =
\begin{cases}
z_c(u), & \operatorname{sign}(z_c(u)) = d_c(u),\\
\eta z_c(u), & \operatorname{otherwise}.
\end{cases}
\end{equation}
Here, $\eta=1$ leaves both polarities unchanged, whereas smaller values increasingly suppress responses whose sign disagrees with the locally dominant one.
The gated response is then obtained by interpolating between the original response and the competitive target:
\begin{equation}
\hat z_c(u)
=
\bigl(1-g_c(u)\bigr)z_c(u)
+
g_c(u)\,\zeta_c\,\bar z_c(u),
\end{equation}
where $\zeta_c$ is a channel-wise normalization factor chosen to preserve the total response energy of channel $c$ in the competitive target. The gate $g_c(u)$ determines how strongly the local competition is applied. In one-sided regions, where one polarity clearly dominates and $g_c(u)$ is small, the original response is left nearly unchanged. When both polarities have non-negligible support within $\mathcal N(u)$, $g_c(u)$ increases and the response is progressively biased toward the locally dominant polarity. The competition is therefore local and label-free, and its purpose is to reduce sign ambiguity before the responses are converted into spike latencies.

\noindent \textbf{Polarity split.} After gating, the two polarities are explicitly represented as two separate non-negative maps
\begin{equation}
r_{c,\sigma}(u) = [\sigma \hat z_c(u)]_+, \quad \sigma \in \{+,-\}.
\end{equation}
Thus, for static inputs, the polarity split maps each signed channel into two non-negative event channels. If the sensory tensor has $C_\Pi$ channels, the split produces $2C_\Pi$ response maps. This representation makes both positive and negative contrast patterns available to STDP as non-negative spike events, ensuring that information carried by either polarity can contribute to local plasticity.

\noindent \textbf{Calibration.}
For each polarity channel $c$ and spatial location $u$, we estimate the extrema from the training split as
\begin{equation}
m_{c,\sigma}(u)
=
\min_{x\in\mathcal D_{\rm train}}
r^{(x)}_{c,\sigma}(u),
\quad\,
M_{c,\sigma}(u)
=
\max_{x\in\mathcal D_{\rm train}}
r^{(x)}_{c,\sigma}(u).
\end{equation}
These values are then used to calibrate the response as:
\begin{equation}
a_{c,\sigma}(u)
=
\frac{
r_{c,\sigma}(u)-m_{c,\sigma}(u)
}{
M_{c,\sigma}(u)-m_{c,\sigma}(u)
},
\end{equation}
whenever $M_{c,\sigma}(u)>m_{c,\sigma}(u)$. If the two extrema coincide, we set $a_{c,\sigma}(u)=0$. The extrema remain fixed after their initial estimation.

The calibrated positive and negative responses are then locally rebalanced within local patches. Let $\mathcal P$ denote the patch containing location $u$, over which polarity statistics and normalization are computed jointly. The rebalanced response is
\begin{equation}
a^\star_{c,\sigma}(u)
=
\lambda_{\mathcal P}\,
\omega_{c,\sigma}(u)\,
a_{c,\sigma}(u),
\end{equation}
where $\omega_{c,\sigma}(u)$ is a deterministic local reweighting factor derived from the relative polarity activity, and $\lambda_{\mathcal P}$ is chosen such that the summed positive and negative response mass within the local patch $\mathcal P$ is preserved. The reweighting reduces excessive positive-polarity dominance and suppresses weak conflicting responses in sufficiently active regions. Depending on the dataset configuration, polarity balancing may also include a local competition between opposite-polarity responses, attenuating the weaker response while preserving the overall local response magnitude. The resulting responses $a^\star_{c,\sigma}$ are used for latency encoding.

\noindent \textbf{Event-based data.}
For event-based inputs, the preceding static-image transformations are replaced by a dedicated preprocessing procedure, called ``EventMap'', that operates directly on the asynchronous event stream, with each sensory sample represented as
\[
\mathcal{E}
=
\left\{
\left(u_i,t_i,\sigma_i\right)
\right\}_{i=1}^{N},
\]
where $u_i=(x_i,y_i)$ denotes the spatial location, $t_i$ the event timestamp, and $\sigma_i\in\{+,-\}$ its polarity. Before constructing the latency representation, isolated events are removed using the following filter, acting as a spatiotemporal denoiser. Specifically, an event at time $t_i$ is discarded if no other event occurs within its one-pixel spatial neighborhood in $[t_i-\delta_t, t_i+\delta_t]$, where $\delta_t$ is a parameter.
The temporal extent of each sample is then normalized between its first and last retained events. Specifically,
\begin{equation}
\tau_i
=
\frac{t_i-t_{\min}}
{t_{\max}-t_{\min}},
\end{equation}
and the normalized interval is partitioned into $B$ uniform temporal bins. Each event $i$ is assigned to as specific bin $b_i$ as follows, 
\begin{equation}
b_i
=
\min
\left\{
B-1,\,
\left\lfloor B\tau_i \right\rfloor
\right\}.
\end{equation}

Separate event-count maps are formed for each position $u_i$, temporal bin $b_i$ and polarity $\sigma$:
\begin{equation}
N_{b,\sigma}(u)
=
\left|
\left\{
i :
u_i=u,\;
\sigma_i=\sigma,\;
b_i=b
\right\}
\right|.
\end{equation}
Thus, temporal order is retained at a coarse resolution through the bin index $b$, while the number of events within each bin and polarity determines the local response strength.

Next, the count maps are logarithmically compressed and normalized with respect to their local spatial mean:
\begin{equation}
q_{b,\sigma}(u)
=
\frac{
\log\!\left(1+N_{b,\sigma}(u)\right)
}{
\operatorname{Avg}_{v\in\mathcal{N}_r(u)}
\log\!\left(1+N_{b,\sigma}(v)\right)
}.
\end{equation}
A sample-wise normalization is subsequently applied across temporal bins, polarities, and spatial locations:
\begin{equation}
a^\star_{b,\sigma}(u)
=
\left[
\frac{
q_{b,\sigma}(u)
}{
\displaystyle
\max_{b',\sigma',v}
q_{b',\sigma'}(v)
}
\right]^{\,\gamma}.
\end{equation}
Zero-valued locations remain inactive. The resulting $2B$ non-negative maps (one per polarity) are passed to the common response-to-latency conversion described below.

Unlike the static-image front end, this event-based transformation does not rely on dataset-level statistics estimated from the training split. In fact, binning, denoising, compression, and normalization are deterministic operations, with normalization performed independently for each sample.

\noindent \textbf{Latency encoding.}
Following the input-specific processing described above, both static and event-based representations are converted into spike times through a common latency-encoding procedure.

To simplify the notation, the non-negative response maps produced by the front end are collected along a common channel index $j$. For static inputs, $j$ indexes the polarity-resolved channels $(c,\sigma)$, with $j=2c$ for positive polarity and $j=2c+1$ otherwise. Instead, for event-based inputs $j$ indexes the polarity-time channels $(b,\sigma)$. In both cases, we denote the resulting response at spatial location $u$ by $a^\star_j(u)$. The final front-end stage converts these response strengths into first-spike latencies.
Responses exceeding the unit range are saturated before the
latency mapping:
\begin{equation}
\tilde a^\star_j(u)=\min\{a^\star_j(u),1\}.
\end{equation}
The normalized spike latency at channel $j$ and spatial location $u$ is then obtained as:
\begin{equation}
L_j(u) =
\begin{cases}
\ell_{\max}
-
(\ell_{\max}-\ell_{\min})\,\tilde a^\star_j(u),
& a^\star_j(u)>\theta_{\mathrm{silent}},\\
\infty,
& \text{otherwise}.
\end{cases}
\end{equation}
Here, $\theta_{\mathrm{silent}}$ denotes the response threshold below which no spike is emitted. In the experiments reported in this paper, we use the normalized latency interval $[\ell_{\min},\ell_{\max}]=[0,1]$. Thus, stronger active responses produce earlier spikes, responses exceeding the unit range are assigned the minimum latency $\ell_{\min}$, and responses below threshold $\theta_{\mathrm{silent}}$ remain inactive.

The front end therefore produces continuous latency maps. For static inputs, continuous latencies are passed directly to \(S_1\), whereas event-based latencies are discretized on a finite temporal grid before entering the backbone. After front-end processing, static grayscale and RGB inputs respectively yield two and six latency maps, whereas event-based inputs yield one latency map for each polarity-time channel. As shown in Fig.~\ref{fig:general_schema}, these maps constitute the output of the \emph{Preprocessing} block and the subsequent input to S1, the first stage of the four-layer spiking convolutional backbone S1-S4. The complete processing sequence is summarized in Algorithm~\ref{alg:visual_front_end}.

\begin{algorithm}[t]
\caption{Label-free early-vision latency encoder}
\label{alg:visual_front_end}
\begin{algorithmic}[1]
\Require Training samples $\mathcal{D}_{\mathrm{train}}$, sample $x$
\Ensure Latency tensor $H^{(0)}=\Phi(x)$

\Statex \textbf{Fit front-end statistics}
\If{static-image input}
    \State Extract patches from $\Pi(\mathcal{D}_{\mathrm{train}})$.
    \State Estimate $\mu,\Sigma$ and compute the local normalization matrix $W=(\Sigma+\epsilon I)^{-1/2}$.
    \State Estimate calibration statistics $\{m_j(u),M_j(u)\}$.
    \State Estimate polarity-balancing normalizers.
\Else
    \State Fix event-map binning and normalization parameters.
\EndIf
\State Freeze all estimated quantities.

\Statex \textbf{Encode sample}
\State $p \gets \Pi(x)$
\If{static-image input}
    \State $z \gets \mathrm{LocalDecorrelate}(p;\mu,W)$
    \State $\hat z \gets \mathrm{SignedContextGate}(z)$
    \State $r \gets \mathrm{PolaritySplit}(\hat z)$
    \State $a \gets \mathrm{Calibrate}(r;\{m_j,M_j\})$
    \State $a^\star \gets \mathrm{BalancePolarities}(a)$
\Else
    \State $a^\star \gets \mathrm{EventMap}(p)$
\EndIf
\State $H^{(0)} \gets \mathrm{LatencyEncode}(a^\star)$
\State \Return $H^{(0)}$
\end{algorithmic}
\end{algorithm}

\subsection{Design Principles for Multi-Depth Temporal Fusion}

In this section, we discuss the general principles that underpin the design of the Multi-Depth Temporal Fusion mechanisms within the backbone of the proposed SNN. Their specific implementation is described in Section~\ref{sec:arch_instantiation}.

We distinguish three latency representations involved in the fusion process: $P$ denotes the preserved shallow representation; $I$ is an intermediate representation obtained by further processing
$P$; and $D$ represents a deeper temporal representation derived from $I$, see Fig.~\ref{fig:general_schema}.

The residual principle adopted here is to preserve $P$ directly, while allowing
the intermediate representation to contribute only through a sparse set of
early events. Specifically, we define the residual contribution $\Delta_{\mathrm{res}}$ as
\begin{equation}
\Delta_{\mathrm{res}}
=
\operatorname{TopK}_{k_{\mathrm{res}}}(I),
\end{equation}
where $\operatorname{TopK}_{k}(X)$ denotes the temporal top-$k$ selection operator, which retains the $k$ earliest finite events of the latency representation $X$ and suppresses the remaining ones. In this way,
deeper processing can enrich the representation without replacing the
preserved shallow code.

Referring to the residual/agreement fusion component in
Fig.~\ref{fig:general_schema}, the inception-like aspect of the architecture arises from combining temporal evidence available at different processing depths. In addition to the sparse intermediate contribution $\Delta_{\mathrm{res}}$, the fusion mechanism retains deep events only when they are temporally consistent with the corresponding intermediate events. For events at the same feature location $i$, this occurs when both representations contain a finite event and their latency difference
does not exceed the agreement tolerance $m_{\mathrm{agree}}$. We therefore
define the agreement candidate as
\begin{equation}
\widetilde{\Delta}_{\mathrm{agree}}(i)
=
\begin{cases}
\min\{I(i),D(i)\},
&
\begin{aligned}
&I(i)<\infty,\quad D(i)<\infty,\\[-2pt]
&|I(i)-D(i)|\leq m_{\mathrm{agree}},
\end{aligned}\\[6pt]
\varnothing,
& \text{otherwise},
\end{cases}
\end{equation}
where $\varnothing$ denotes silence. When the agreement condition is satisfied, the earliest of the two events $I$ and $D$ is retained; otherwise, the corresponding location remains silent.

\noindent The agreement contribution is then further sparsified as
\begin{equation}
\Delta_{\mathrm{agree}}
=
\operatorname{TopK}_{k_{\mathrm{agree}}}
\left(\widetilde{\Delta}_{\mathrm{agree}}\right).
\label{delta_agree}
\end{equation}
In this way, isolated or temporally inconsistent deep events are suppressed, while agreement-gated events indicate that deeper processing has produced temporal evidence consistent with the intermediate representation.

The final fusion combines the representation $P$ with the sparse intermediate $\Delta_{\mathrm{res}}$ and agreement-based $\Delta_{\mathrm{agree}}$ contributions, adding sparse and temporally validated evidence without replacing (but preserving) the  early latency code $P$.

\subsection{Instantiation of the Multi-Depth Temporal Fusion Mechanism}
\label{sec:arch_instantiation}

Throughout this section, $H^{(\ell)}$ denotes the latency representation exported after the corresponding processing stage, with $H^{(0)}$ denoting the encoded input to the backbone. The four-stage hierarchy is defined as
\begin{equation}
\begin{aligned}
H^{(1)} &= C_1\!\left(S_1(H^{(0)})\right),\\
H^{(2)} &= C_2\!\left(S_2(H^{(1)})\right),\\
H^{(3)} &= S_3(H^{(2)}),\\
H^{(4)} &= S_4(H^{(3)}).
\end{aligned}
\end{equation}
We reserve $H$ without a stage superscript for the final fused representation.

Referring to Fig.~\ref{fig:general_schema}, the encoded input $H^{(0)}$ is first processed by the two early stages $S_1/C_1$ and $S_2/C_2$, where the spiking layers learn local temporal features and the corresponding $C_1$ and $C_2$ operations perform deterministic min-latency pooling and export the resulting latency representations. The deeper path then processes $H^{(2)}$ through $S_3$ and $S_4$ stages, producing $H^{(4)}$, which provides the deep representation used for temporal agreement. Note that $H^{(3)}$ is an intermediate representation of this deep path and is not directly included in the final code.

Referring to the general construction introduced in the previous section, $H^{(1)}$, $H^{(2)}$, and $H^{(4)}$ correspond respectively to the preserved ($P$), intermediate ($I$), and deep ($D$) representations. Hence, $\Delta_{\mathrm{res}}$ is obtained from $H^{(2)}$, while $\Delta_{\mathrm{agree}}$ is computed from the temporal agreement between $H^{(2)}$ and $H^{(4)}$. The fused representation passed to the classifier is therefore
\begin{equation}
H =
\left[
H^{(1)},\;
\Delta_{\mathrm{res}},\;
\Delta_{\mathrm{agree}}
\right].
\label{eq:backbone_output}
\end{equation}

This is the residual/inception-like fusion used in the proposed model. It is residual because the early $H^{(1)}$ code remains directly available to the classifier, while subsequent stages contribute only sparse additional evidence. Its inception-like character arises from combining representations obtained at different processing depths, with the deepest events retained only when they are temporally consistent with the intermediate code.

\section{Local Learning Framework}
\label{sec:learning_framework}

This section defines the local plasticity rules and the training schedule for the backbone and the readout layers.

\subsection{Learning Rules}
\label{sec:rstdp_rules}

Synaptic adaptation is local at the level of individual connections. The convolutional backbone and the readout share the same causal dependence on the relative first-spike times of pre- and postsynaptic units, but use different weight-dependent update functions. We write the local update in the general form 
\begin{equation}
\Delta w_{ij}
=
\lambda_j
\begin{cases}
A^+ F_+(w_{ij}), & t_i \leq t_j,\\
-A^- F_-(w_{ij}), & t_i > t_j,
\end{cases}
\label{eq:stdp_general}
\end{equation}
where $A^+,A^->0$ denote the potentiation and depression amplitudes, respectively, and $\lambda_j$ is a postsynaptic modulation factor.

For the convolutional STDP layers, the update magnitude depends exponentially on the current synaptic weight:
\begin{equation}
F_+^{\mathrm{conv}}(w)=e^{-\beta w},
\qquad
F_-^{\mathrm{conv}}(w)=e^{\beta(w-1)}.
\label{eq:stdp_conv_weight}
\end{equation}
Potentiation therefore decreases toward the upper weight bound, whereas depression decreases toward the lower bound. Updated weights are clipped to the admissible interval.

The R-STDP readout instead uses an additive update,
\begin{equation}
F_+^{\mathrm{readout}}(w)
=
F_-^{\mathrm{readout}}(w)
=
1,
\label{eq:stdp_readout_weight}
\end{equation}
so that the update magnitude does not directly depend on the current synaptic weight. Readout stability is instead enforced through weight clipping and fixed per-neuron $\ell_1$ normalization. For unsupervised STDP layers, $\lambda_j$ is only determined by local winner selection and layer-specific learning-rate scheduling. These layers therefore learn recurring temporal patterns without using labels.

Before the readout, the fused latency tensor $H$ is flattened into a vector of presynaptic spike-time features. This operation changes only the tensor shape: each entry remains a first-spike latency or a silent feature. The readout is therefore implemented as a fully connected spiking decision layer: each output neuron receives the flattened temporal representation and emits one class-prototype latency. The readout follows the same pre/post spike-timing criterion introduced in \eq{eq:stdp_general}, with the additive choice $F_+^{\mathrm{readout}}(w)=F_-^{\mathrm{readout}}(w)=1$, while $\lambda_j$ is determined by a class-level {\it reward signal}. Each class is represented by a small population of readout neurons; a technique known as ``population coding''. Given an input sample with label $y$, the readout produces one latency per output neuron, and the representative latency $\tau_c$ of class $c$ is the earliest finite latency among its prototypes. Therefore, the predicted class is 
\begin{equation}
\hat y = \myargmin{c} \tau_c.
\end{equation}

\noindent \textbf{Reward-modulated learning.} Firstly, we define a sample-dependent temporal corridor around the current class responses:
\begin{equation}
\bar{\tau} =
\frac{1}{|\mathcal{F}|}\sum_{c\in\mathcal{F}}\tau_c,
\qquad
\tau_y^\star = \bar{\tau}-\frac{m}{2},
\qquad
\tau_{\neg y}^\star = \bar{\tau}+\frac{m}{2},
\end{equation}
where $\mathcal{F}$ is the set of classes with finite readout activity and $m$ is a temporal margin. The {\it target class} $c=y$ is encouraged to fire {\it before} the corridor center, while {\it non-target classes} are encouraged to fire {\it after} it. The modulation that is applied to the target class is proportional to the violation of this temporal objective. Hence, for each output neuron $j$ belonging to the target class (set $\mathcal{P}_y$), it holds
\begin{equation}
\lambda_j^+ =
\operatorname{clip}
\left(
\beta_+
\frac{[\tau_j-\tau_y^\star]_+}{\tau_{\max}},
0,\lambda_{\max}
\right),
\qquad j\in\mathcal{P}_y,
\label{eq:reward_modualtion_target}
\end{equation}
where $\tau_{\max}$ is the maximum normalized latency scale, $\beta_+$ is a target-side gain, and $\lambda_{\max}$ limits the magnitude of the update. According to \eq{eq:reward_modualtion_target}, a target neuron that already fires early receives little or no update; a target neuron that fires too late receives a stronger rewarded STDP update.

For each non-target class $c \neq y$, we compute the distance of the corresponding output from the upper margin $\tau_{\neg y}^\star$
\begin{equation}
v_c = [\tau_{\neg y}^\star-\tau_c]_+, \qquad c \neq y.
\end{equation}
Punishment is sparse. Rather than depressing every non-target class, the proposed rule selects the $K$ classes that violate the temporal margin the most. This sparse update ensures stability, by avoiding unnecessary depression of non-competing classes and concentrating plasticity on the outputs that currently interfere the most with the target decision. Accordingly, only the $K$ largest violations are treated as hard negatives (set $\mathcal{P}_{\mathrm{hn}}$). For each selected non-target class, readout neurons $j \in \mathcal{P}_{\mathrm{hn}}$ are updated with anti-STDP:
\begin{equation}
\lambda_j^- =
-\operatorname{clip}
\left(
\beta_-
\frac{v_{c(j)}}{\tau_{\max}},
0,\lambda_{\max}
\right),
\qquad j\in\mathcal{P}_{\mathrm{hn}},
\label{eq:anti-stdp-rule}
\end{equation}
where $c(j)$ is the class of neuron $j$ and $\mathcal{P}_{\mathrm{hn}}$ is the hard-negative prototype set. When multiple prototypes within the same class are updated, they are ranked by increasing output latency, with decreasing membrane potential used to break equal-latency ties. Using multiple prototypes allows each class to be represented by different spike-time response patterns rather than by a single readout neuron, while updating only the most relevant prototypes limits the number of synapses modified for each sample. The sign of $\lambda_j$ selects reward or punishment, while the synaptic update itself remains timing-based. Positive modulation \eqref{eq:reward_modualtion_target} applies the rewarded STDP rule to target prototypes; negative modulation \eqref{eq:anti-stdp-rule} applies the anti-STDP rule to confusing non-target prototypes. For correctly classified samples, the same margin-based updates are retained, but their modulation is reduced by a dataset-specific scaling factor. If no output neuron fires, output latencies are assigned the maximum latency and membrane potential is used to resolve the resulting tie; in this low-activity condition, anti-STDP \eqref{eq:anti-stdp-rule} is suppressed, and only target-side update \eqref{eq:reward_modualtion_target} is applied. This readout training approach remains within the R-STDP family, replacing the purely binary reward of pure-STDP with a local temporal objective focused on the classes that actually compete for the decision.

The mechanism amounts to a temporal competition between class populations: target-class prototypes are reinforced when they fire too late, whereas the most competitive non-target prototypes are punished when they fire too early. The remaining details, including clipping of modulation magnitudes, rank-based decay across prototypes, and silent-state guards, are stabilization mechanisms used to keep the readout in an active but sparse firing regime.

\subsection{Training Schedule}
\label{subsec:training_schedule}

Training follows the model structure. The network is trained layer by layer, with no end-to-end backpropagation. Each trainable layer is exposed to the spike-time representation produced by the preceding layers, updates only the synapses assigned to that layer, and is then frozen before the next layer is trained. This schedule restricts each backbone layer to local spike-timing information, while task-dependent modulation is introduced only at the readout.

The visual front end is configured before training the backbone. For static inputs, the patch mean $\mu$, covariance $\Sigma$, local covariance-normalization matrix $W$, calibration ranges, and polarity-balancing normalizers are estimated once from the training split and then kept fixed for training, validation, and test phases. For event-based inputs, the front end instead uses fixed denoising, temporal binning, and normalization operations applied independently to each sample, without using dataset-level statistics. No label information is used in either case and the front end acts as a deterministic encoder, which makes it possible to reuse the same latency maps across all backbone and readout variants.

The convolutional backbone is trained layerwise. The spiking layers $S_1$ and $S_2$ learn recurring local temporal patterns through unsupervised STDP, while $C_1$ and $C_2$ deterministically export the corresponding latency representations. Once a spiking layer has been trained, it is frozen and its output representation is cached for the following layers. The deeper section of the backbone processes the intermediate representation $I=H^{(2)}$ through $S_3$ and $S_4$, producing the deep representation $D=H^{(4)}$ used for temporal agreement. The final representation is then constructed deterministically by preserving $P=H^{(1)}$, extracting the sparse residual contribution $\Delta_{\mathrm{res}}$ from $I$, and adding the agreement contribution $\Delta_{\mathrm{agree}}$ obtained from temporally consistent events in $I$ and $D$. These three contributions are concatenated to form the fused latency representation $H=[P,\Delta_{\mathrm{res}},\Delta_{\mathrm{agree}}]$ used by the readout classifier, see \eq{eq:backbone_output}.

The readout layer is trained after the backbone has been fixed. During readout training, each input sample is forwarded through the frozen backbone, output latencies are computed, and the R-STDP update described in Section~\ref{sec:rstdp_rules} is applied immediately (samplewise). This preserves the online character of the learning rule and keeps the supervision signal local to the final decision stage.

Table~\ref{tab:training_schedule} summarizes the resulting schedule. Exact numerical hyperparameters, such as epoch counts, thresholds, top-$k$ values, temporal margins, and modulation caps, are treated as experimental settings and specified in Appendix~\ref{app:reproducibility}.

\begin{table}[t]
    \centering
    \scriptsize
    \renewcommand{\arraystretch}{1.18}
    \caption{Training schedule of the proposed locally trained pipeline.}
    \label{tab:training_schedule}
    \begin{tabular}{p{0.18\textwidth}p{0.22\textwidth}p{0.42\textwidth}}
        \toprule
        \textbf{Stage} & \textbf{Plasticity} & \textbf{Role in the pipeline} \\
        \midrule
        Visual front end
        & None; deterministic preprocessing
        & Converts static images or event streams into label-free latency representations. Static-input statistics are estimated once from the training split, whereas event-based normalization is applied independently to each sample. \\
        
        Early convolutional stages
        & Unsupervised STDP in $S_1$-$S_2$; deterministic $C_1$-$C_2$ export
        & To learn early temporal features and produce the preserved representation $P=H^{(1)}$ and the intermediate representation $I=H^{(2)}$; each trained spiking stage is frozen before subsequent stages are fitted. \\
        
        Deep convolutional path
        & Unsupervised STDP in $S_3$-$S_4$
        & Further processes $I$ to produce the deep representation $D=H^{(4)}$, used together with $I$ by the temporal agreement mechanism. \\
        
        Sparse Multi-Depth Temporal Fusion
        & None; deterministic selection and agreement
        & Preserves $P$, extracts the sparse residual contribution $\Delta_{\mathrm{res}}$ from $I$, and adds the agreement contribution $\Delta_{\mathrm{agree}}$ obtained from temporally consistent events in $I$ and $D$. \\
        
        Multi-prototype readout
        & Reward-modulated STDP
        & Learns class decisions from the frozen fused representation $H=[P,\Delta_{\mathrm{res}},\Delta_{\mathrm{agree}}]$ using target-side reinforcement, hard-negative anti-STDP, and multiple prototypes per class. \\
        \bottomrule
    \end{tabular}
\end{table}

This layer-wise organization also determines the ablation study presented in Section~\ref{sec:experiments}. Its core components are the latency front end, the locally trained STDP backbone, the sparse MDTF module, and the multi-prototype R-STDP readout layer. Front-end components are evaluated separately through dedicated controls, while readout stabilization mechanisms are kept fixed within each reported operating point. This separation makes it possible to distinguish improvements due to a better input latency code, the residual and inception-based temporal routing in the backbone, or reward-modulated decision learning.

\section{Experiments and Results}
\label{sec:experiments}

The experimental evaluation is designed to test the central hypothesis of this work: deep architectural structure can be made useful in an SNN trained by local synaptic plasticity. All reported models use the layer-wise training protocol described in Section~\ref{sec:learning_framework}. The preprocessing transform is estimated from the training data split without using labels, the convolutional spiking backbone is trained by local STDP, and the final readout layer is trained on frozen spike-time features using reward-modulated STDP. Class labels are therefore only used by the reward signal of the readout; they are not used to train the backbone.

For reproducibility, the settings for all training parameters are reported in Appendix~\ref{app:reproducibility} for all the considered datasets.

\subsection{Datasets and Performance Metrics}

\noindent \textbf{Benchmarks.} The proposed design is evaluated on static and event-based tasks. The considered datasets have an increasing complexity. MNIST (static, grayscale digits) is the simplest, providing a low-variability setting. Fashion-MNIST (static, grayscale apparel) preserves the same spatial format but introduces a substantially more heterogeneous shape distribution across input samples. CIFAR-10 (static, RGB images) features more complex input images with colors. N-MNIST (dynamic, event-based) features event-based signals which serve to evaluate the event-based version of the front end.

\noindent \textbf{Performance metrics.} The primary metric is {\it test-set accuracy}. For the R-STDP readout, however, accuracy is interpreted jointly with firing stability. The epoch used for reporting test accuracy is selected according to validation performance. Test-set trajectories are reported only to assess the stability of the readout dynamics over training and are not used for model selection.

\subsection{Main Results}
\label{sec:main_accuracy_results}

The numerical results show a consistent pattern across the four benchmarks, see Table~\ref{tab:main_results}. On MNIST, the locally trained representation/readout pipeline reaches the high-accuracy regime expected for a low-variability digit task, indicating that the proposed latency code does not sacrifice performance in this simple setting. Fashion-MNIST introduces substantially more shape variability across input samples; the corresponding results show that the proposed SNN is as well successful on a more heterogeneous recognition problem. CIFAR-10 is more demanding because the front end in this case must preserve weak color and texture evidence, impacting the accuracy. The event-based N-MNIST is a dynamic input: also in this case, once asynchronous polarity events are converted into a locally normalized event-map latency representation, the proposed four-layer backbone and R-STDP readout remain effective.

\begin{table}[t]
    \centering
    \footnotesize
    \renewcommand{\arraystretch}{1.15}
    \caption{Main classification results and frozen-code statistics. Accuracy is reported in percent on the test split for validation-selected operating points; multi-seed rows report mean $\pm$ standard deviation over readout seeds. Code dimension, mean active events, and active density are measured on the sparse latency representation delivered to the readout. All rows use the same four-layer locally trained residual/agreement spiking backbone and a reward-modulated temporal R-STDP readout.}
    \label{tab:main_results}
    \begin{tabular}{llrrrr}
        \toprule
        \textbf{Dataset} & \textbf{Input structure} & \textbf{Code dim.} & \textbf{Events/sample} & \textbf{Density} & \textbf{Accuracy (\%)} \\
        \midrule
        MNIST & $28{\times}28$ grayscale digits & $17\,408$ & $994.5$ & $5.71\%$ & $\meanstd{96.6}{0.6}$\\
        Fashion-MNIST & $28{\times}28$ grayscale apparel & $17\,408$ & $1\,182.3$ & $6.79\%$ & $\meanstd{86.3}{0.7}$ \\
        CIFAR-10 & $32{\times}32$ RGB natural images & $24\,704$ & $3\,100.6$ & $12.6\%$ & $\meanstd{62.5}{0.4}$ \\
        N-MNIST & $34{\times}34$ polarity event streams & $24\,704$ & $1\,992.7$ & $8.07\%$ & $\meanstd{95.1}{0.6}$ \\
        \bottomrule
    \end{tabular}
\end{table}

\begin{figure}[t]
    \centering
    \includegraphics[width=0.93\textwidth]{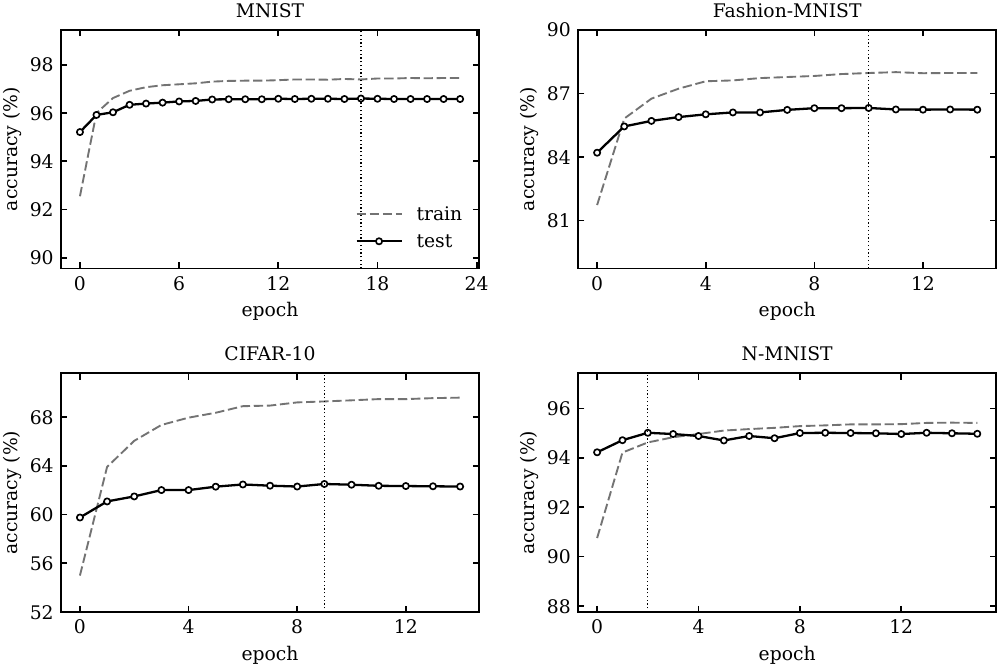}
    \caption{Training dynamics of the reward-modulated readout on the main datasets. Solid black curves report test accuracy, dashed gray curves report training accuracy, and the vertical dotted line marks the best validation epoch. The panels show stable readout regimes, with test accuracy remaining close to its best value over the final training phase.}
    \label{fig:readout_dynamics}
\end{figure}

The code statistics in Table~\ref{tab:main_results} show that these accuracies are obtained from {\it sparse} frozen representations. As the input structure becomes richer, the exported code uses more active latency events, but remains sparse relative to its full dimensionality. Fig.~\ref{fig:readout_dynamics} provides a stability check on the readout dynamics: test accuracy remains close to its best value over the final training phase, with no isolated transient peak.

\subsection{Front-End Configuration Ablations}
\label{sec:frontend_configurations}

The input encoder implements a sequence of deterministic operations. For each, in this section we compare the full front end with alternative configurations in which selected processing mechanisms are removed or replaced. The backbone and readout retain the same four-layer architecture and learning framework within each dataset.

\begin{table}[t]
    \centering
    \footnotesize
    \renewcommand{\arraystretch}{1.15}
    \caption{Front-end component configurations. Table entries report test accuracy in percent on the full train/test protocol, using the same four-layer backbone architecture and R-STDP readout framework.}
    \label{tab:frontend_component_controls}
    \begin{tabular}{lllr}
        \toprule
        \textbf{Dataset} & \textbf{Front-end configuration} & \textbf{Main configuration change} & \textbf{Accuracy} \\
        \midrule
        MNIST & simple latency & local decorrelation and polarity processing & $9.80$ \\
        MNIST & full without signed context & signed-context gate & $95.41$ \\
        MNIST & full without polarity balancing & local polarity-balance gates & $95.02$ \\
        MNIST & full static front end & - & $96.68$ \\
        \midrule
        Fashion-MNIST & simple latency & local decorrelation and polarity processing & $10.00$ \\
        Fashion-MNIST & full without signed context & signed-context gate & $86.23$ \\
        Fashion-MNIST & full without polarity balancing & local polarity-balance gates & $86.24$ \\
        Fashion-MNIST & full static front end & - & $86.31$ \\
        \midrule
        CIFAR-10 & RGB latency & local decorrelation, polarity split, calibration & $9.8$ \\
        CIFAR-10 & decorrelation + polarity + calibration & local polarity balancing & $60.48$ \\
        CIFAR-10 & channel-peak calibration & dataset-level response calibration & $58.12$ \\
        CIFAR-10 & full static front end & - & $62.50$ \\
        \midrule
        N-MNIST & raw event frames & event-map construction & $93.16$ \\
        N-MNIST & full map without denoising & event denoising filter & $93.49$ \\
        N-MNIST & full map without log compression & log compression & $93.51$ \\
        N-MNIST & full event-map & - & $95.01$ \\
        \bottomrule
    \end{tabular}
\end{table}

The results of this ablation study are reported in Table~\ref{tab:frontend_component_controls}. 
Direct latency coding from raw intensity (termed ``simple latency'' or ``RGB latency'', depending on the dataset) is close to chance on all static benchmarks, showing that a usable temporal code {\it does not emerge from raw intensity ordering alone}. On MNIST, removing signed-context or local polarity-balance processing produces measurable losses, whereas on Fashion-MNIST all configurations perform close to the ``full front end''. On CIFAR-10, local decorrelation combined with polarity split and response calibration achieves most of the full static-front-end performance, whereas the alternative channel-peak configuration performs less effectively. On the N-MNIST event-based dataset, raw event frames already provide a strong event-based code, achieving accuracies higher than $93\%$, but the full event-map representation returns the best results among the tested configurations. In this case, removing denoising or log compression gives a modest but consistent loss.

\subsection{Comparison against State-of-the-Art STDP/R-STDP Baselines}
\label{sec:comparison_with_sota}

Next, we compare the proposed architecture with a local STDP/R-STDP baseline that shares the same broad processing logic: temporal coding, unsupervised STDP feature learning, and a final spiking decision stage trained through reward-modulated plasticity. The selected baseline follows the convolutional STDP/R-STDP scheme of~\cite{mozafari_bio-inspired_2019}, which is particularly close to our setting because classification is performed directly in the spiking domain without an external classifier. Other local spike-timing approaches follow related but different learning schemes, for example by combining unsupervised STDP features with a separate classifier \cite{kheradpisheh_stdp-based_2018, falez_unsupervised_2019} or by using directly supervised synaptic updates \cite{hao_biologically_2020, goupy_paired_2024} instead of reward-modulated plasticity. The comparison is therefore intended to contextualize the effect of the proposed representation and routing mechanisms within a closely matched local STDP/R-STDP setting.

Table~\ref{tab:local_baseline_comparison} presents the experimental results of this comparison. On MNIST, the baseline already provides high-accuracy, and the proposed model does not present noticeable improvements. The advantage offered by the proposed architecture appears on more complex datasets where the early latency code alone is no longer sufficient. On Fashion-MNIST and CIFAR-10, our method gives major gains over the baseline. The baseline underperforms on N-MNIST; this is because it does not account for a front end layer to express dynamic inputs in a latency format that is suitable for further SNN processing. This result highlights the importance of transforming asynchronous event streams into a spike-time representation that is compatible with the backbone.

\begin{table}[t]
    \centering
    \footnotesize
    \renewcommand{\arraystretch}{1.15}
    \caption{Comparison with a local STDP/R-STDP baseline from~\cite{mozafari_bio-inspired_2019}. Accuracies are reported in percent with $95\%$ Wilson confidence intervals over the test split. The $p$-values are Holm-corrected two-proportion tests at the aggregate accuracy level. Very small corrected values are clipped at $10^{-16}$ for readability.}
    \label{tab:local_baseline_comparison}
    \begin{tabular}{lrrrr}
        \toprule
        \textbf{Dataset} & \textbf{Local STDP/R-STDP baseline} & \textbf{Proposed full model} & \textbf{$\Delta$ pp} & \textbf{$p_{\mathrm{Holm}}$} \\
        \midrule
        MNIST & \accci{97.0}{96.6}{97.3} & \accci{96.7}{96.3}{97.1} & $-0.3$ & $0.18$ \\
        Fashion-MNIST & \accci{68.2}{67.5}{69.1} & \accci{86.3}{85.6}{87.0} & $+18.2$ & $<10^{-16}$ \\
        CIFAR-10 & \accci{33.3}{32.4}{34.2} & \accci{62.5}{61.6}{63.5} & $+29.2$ & $<10^{-16}$ \\
        N-MNIST & \accci{22.0^{\dagger}}{21.2}{22.8} & \accci{95.0}{94.6}{95.5} & $+73.0$ & $<10^{-16}$ \\
        \bottomrule
    \end{tabular}

    \vspace{0.25em}
    \raggedright
    \scriptsize $^{\dagger}$Direct-transfer diagnostic: the local STDP/R-STDP baseline is applied to N-MNIST event streams without the event-map preprocessing used by the proposed model.
\end{table}

\subsection{Data Efficiency}
\label{sec:data_efficiency}

The data-efficiency experiments retrain the layer-wise pipeline at progressively larger training budgets, while keeping the readout training protocol fixed. Reducing the number of examples affects both the unsupervised STDP layers and the reward-modulated readout, so the experiment probes the full local-learning pipeline, not only the final classifier. Table~\ref{tab:data_efficiency} reports the resulting accuracy results. 

\begin{table}[t]
    \centering
    \footnotesize
    \renewcommand{\arraystretch}{1.15}
    \caption{Data-efficiency for the proposed pipeline. Entries are test accuracies in percent {\it vs} number of training examples.}
    \label{tab:data_efficiency}
    \begin{tabular}{lrrrrrrr}
        \toprule
        \textbf{Dataset} & \textbf{100} & \textbf{500} & \textbf{1k} & \textbf{5k} & \textbf{10k} & \textbf{20k} & \textbf{30k} \\
        \midrule
        MNIST & $41.45$ & $78.41$ & $84.08$ & $91.90$ & $93.17$ & $94.28$ & $94.93$ \\
        Fashion-MNIST & $49.59$ & $71.20$ & $74.42$ & $80.73$ & $82.45$ & $84.37$ & $85.02$ \\
        CIFAR-10 & $12.57$ & $34.98$ & $41.66$ & $53.54$ & $56.40$ & $59.31$ & $61.16$ \\
        N-MNIST & $27.67$ & $51.43$ & $68.11$ & $88.50$ & $90.78$ & $92.44$ & $93.29$ \\
        \bottomrule
    \end{tabular}
\end{table}

\begin{figure}[t]
    \centering
    \subfloat[Accuracy scaling.]{
        \includegraphics[width=0.47\textwidth]{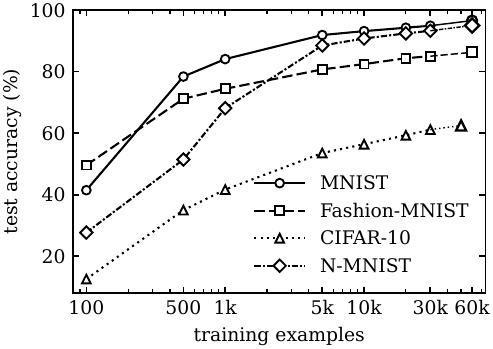}
    }
    \hfill
    \subfloat[Train-test gap.]{
        \includegraphics[width=0.47\textwidth]{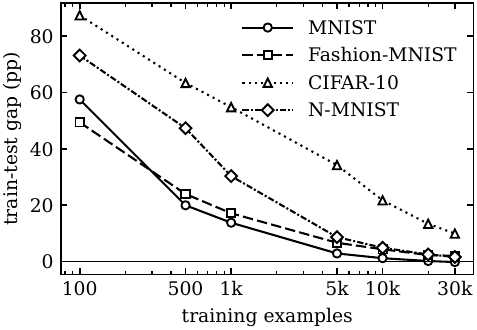}
    }
    \caption{Data-efficiency behaviour of the locally trained pipeline. The figure uses line pattern and marker shape rather than color: circles denote MNIST, squares denote Fashion-MNIST, triangles denote CIFAR-10, and diamonds denote N-MNIST.}
    \label{fig:data_efficiency_scaling}
\end{figure}

Fig.~\ref{fig:data_efficiency_scaling} shows the same experiment as a scaling curve. Accuracy increases monotonically on all four datasets, indicating that both the STDP backbone and the R-STDP readout benefit from additional training samples. MNIST and Fashion-MNIST show the expected early-saturation pattern for grayscale recognition, with Fashion-MNIST reaching a lower ceiling because of its larger intra-class variability. CIFAR-10 remains the most data-demanding static benchmark, while N-MNIST improves sharply once the event-map representation leaves the few-sample regime. The monotonic scaling indicates that the locally trained pipeline continues to benefit from additional training data over the tested range.

\subsection{Spike-Budget Pareto}
\label{sec:spike_budget_Pareto_analysis}

The sparse latency representation at the output of the backbone allows the final readout layer to be evaluated under an explicit {\it activity budget}. To do so, for each input sample we retain the top-scoring sparse spike features at the output of the backbone and retrain the readout on the truncated code. The score is induced by latency, so earlier spikes correspond to stronger retained evidence. This experiment does not change the trained backbone. Rather, it probes how much of its final representation is required by the readout layer to provide a certain performance level. The truncation amounts to removing weak or late events from the backbone output. Because the datasets have different native activity levels, the retained budget is expressed as a percentage of the mean full-code event count on the test split.

\begin{figure}[t]
    \centering
    \subfloat[Spike-budget Pareto.]{
        \includegraphics[width=0.47\textwidth]{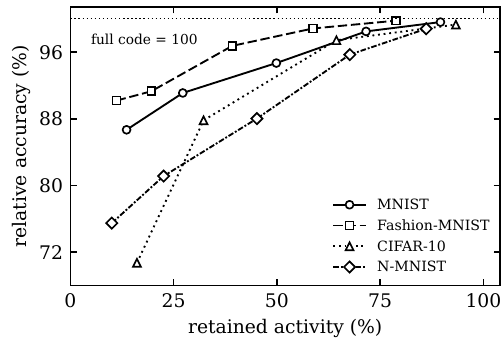}
    }
    \hfill
    \subfloat[Operating-point budget.]{
        \includegraphics[width=0.47\textwidth]{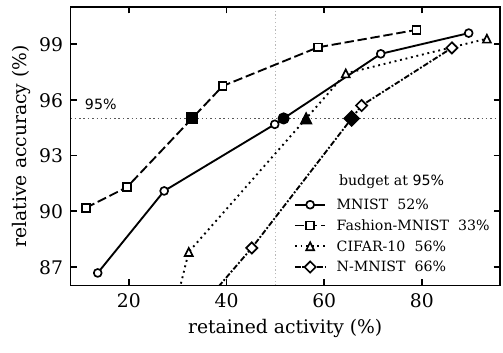}
    }
    \caption{Spike-budget Pareto and event-operation proxy for the final sparse latency code. (a) Test accuracy relative to the matched full-code readout after retaining only the strongest latency events in each sample and retraining the R-STDP readout with the backbone fixed. The horizontal dotted line marks the untruncated code normalized to $100\%$. (b) Zoom on the high-accuracy operating regime. Filled markers indicate the interpolated retained activity required to preserve $95\%$ of full-code accuracy, yielding a compact proxy for the event-operation budget of the readout code.}
    \label{fig:spike_budget_pareto}
\end{figure}

Fig.~\ref{fig:spike_budget_pareto} shows a gradual loss of accuracy as activity is reduced. The simpler static benchmarks retain most of their full-code performance at relatively small budgets, whereas CIFAR-10 and N-MNIST require more retained events to reach the high-accuracy regime. This indicates that the fused sparse code contains a concentrated subset of high-value early evidence, and that removing later or weaker events is a suitable means to control the tradeoff between maximum accuracy and readout activity.

\subsection{Residual and Agreement Ablations}
\label{sec:residual_agreement_ablations}

The routing ablation clarifies the role of depth in the proposed architecture. We expect representations obtained at increasing processing depth to be most effective when they provide sparse corrections to the preserved representation \(P=H^{(1)}\), not as standalone replacements. To check this, we compare residual readouts that preserve \(P\) and progressively incorporate contributions derived from the intermediate representation \(I=H^{(2)}\) and the deep representation \(D=H^{(4)}\). For the diagnostic route without temporal agreement, we denote the directly selected deep contribution by
\begin{equation}
\Delta_D=\operatorname{TopK}_{k_D}(D).
\end{equation}
This quantity is used only in the ablation: the proposed model instead retains the agreement-filtered contribution \(\Delta_{\mathrm{agree}}\), see \eqref{delta_agree}. All variants use the same trained frozen backbone and the same temporal R-STDP readout protocol; only the combination of backbone representation presented to the readout is changed.

\begin{figure}[t]
    \centering
    \subfloat[Residual routing ablation.]{
        \includegraphics[width=0.47\textwidth]{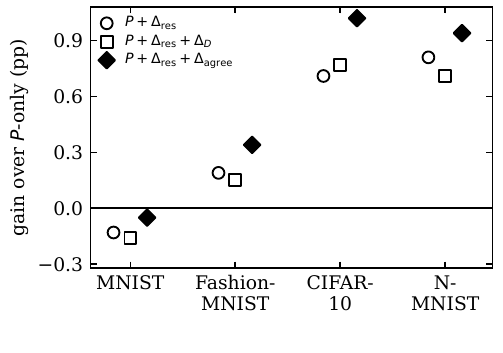}
        \label{fig:routing_residual_ablation}
    }
    \hfill
    \subfloat[Temporal evidence preservation.]{
        \includegraphics[width=0.47\textwidth]{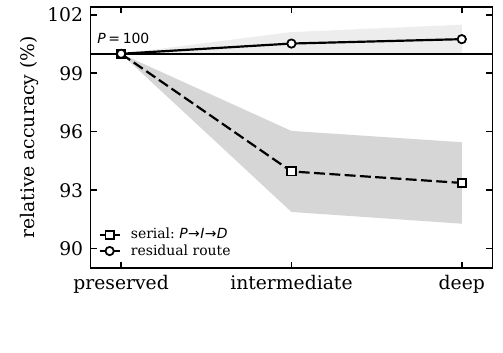}
        \label{fig:temporal_evidence_preservation}
    }
    \caption{Residual routing and temporal-evidence preservation. (a) Percentage-point changes relative to the preserved representation \(P=H^{(1)}\) for the residual route \([P,\Delta_{\mathrm{res}}]\), its direct deep extension \([P,\Delta_{\mathrm{res}},\Delta_D]\), and the proposed agreement-filtered code \([P,\Delta_{\mathrm{res}},\Delta_{\mathrm{agree}}]\). (b) Mean retained accuracy across the four datasets, normalized by the \(P\)-only route; shaded envelopes indicate across-dataset dispersion. At the three processing depths, the serial curve evaluates \(P\), \(I\), and \(D\), whereas the residual curve evaluates \(P\), \([P,\Delta_{\mathrm{res}}]\), and \([P,\Delta_{\mathrm{res}},\Delta_{\mathrm{agree}}]\), respectively.}
    \label{fig:routing_temporal_evidence}
\end{figure}

Fig.~\ref{fig:routing_residual_ablation} shows that the preserved representation \(P\) remains the dominant evidence stream, as expected. However, additional contributions from later representations $I$ and $D$ provide selective refinements, especially on the higher-variability datasets. On MNIST, where \(P\) already provides a highly discriminative code, incorporating the additional sparse contributions produces little or no change in accuracy. The scaling analysis in Fig.~\ref{fig:temporal_evidence_preservation} further characterizes the role of depth: serial replacement evaluates $P$ (preserved), $I$ (intermediate) and $D$ (deep) in isolation, losing accuracy when \(P\) is discarded, and showing that $I$ and $D$ alone are insufficient. Residual routing keeps \(P\) available by incorporating sparse contributions derived from \(I\) and then from \(D\), confirming that their integration leads to an increase in the accuracy.

\subsection{Decision-Level Confusion Repair}
\label{sec:decision_analysis}

To analyze how the fused representation changes individual decisions, we compare the $P$-only readout with the full fused representation $H=[P,\Delta_{\mathrm{res}},\Delta_{\mathrm{agree}}]$ on the same test samples.
Each input sample is assigned to one of three transition types: a {\it repaired error}, when the prediction from $P$ is wrong and the prediction from $H$ is correct; a {\it new error}, when the prediction from $P$ is correct and that from $H$ is wrong; or a {\it changed but unresolved error}, when both predictions are wrong but the predicted class changes. This analysis separates genuine corrections from newly introduced errors and changes among already incorrect predictions.

\begin{figure}[t]
    \centering
    \subfloat[Decision transitions $P\rightarrow H$.]{
        \includegraphics[width=0.47\textwidth]{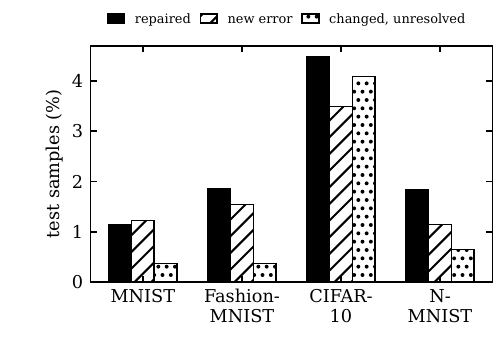}
        \label{fig:confusion_repair_transitions}
    }
    \hfill
    \subfloat[Fate of $P$-only errors.]{
        \includegraphics[width=0.47\textwidth]{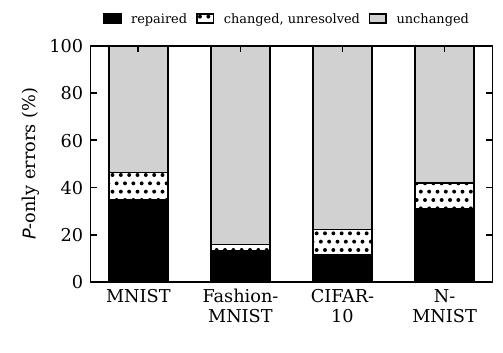}
        \label{fig:confusion_repair_errors}
    }
    \caption{Decision transitions induced by the full fused representation
    $H=[P,\Delta_{\mathrm{res}},\Delta_{\mathrm{agree}}]$ relative to the
    $P$-only readout.
    (a) Fraction of test samples corresponding to repaired errors
    ($P$ wrong, $H$ correct), new errors ($P$ correct, $H$ wrong), and
    changed but unresolved errors (both predictions wrong, but with different
    predicted classes).
    (b) Fate of the errors produced by the $P$-only readout, partitioned into
    repaired errors, changed but unresolved errors, and errors whose predicted
    class remains unchanged.}
    \label{fig:confusion_repair}
\end{figure}

Fig.~\ref{fig:confusion_repair} shows that the full representation only modifies a limited fraction of the test decisions. On Fashion-MNIST, CIFAR-10, and N-MNIST, repaired errors outnumber newly introduced errors, so the additional contributions provide a positive net effect on classification accuracy. CIFAR-10 shows the largest amount of decision change: $H$ repairs a visible subset of the errors made from $P$, while another subset changes predicted class but remains incorrect. MNIST behaves differently because the $P$-only representation is already close to saturation, and the additional contributions produce little net change. Fig.~\ref{fig:confusion_repair_errors} shows the split between errors types: unchanged, changed plus unresolved and repaired. Most errors made from $P$ remain unchanged, whereas $\Delta_{\mathrm{res}}$ and $\Delta_{\mathrm{agree}}$ affect a dataset-dependent subset of the remaining decisions.

\subsection{Interpretation of Representation and Routing}
\label{sec:results_interpretation}

The front-end ablations show that the representation outputted by the front end is a major determinant of performance. On the static datasets, direct latency coding from raw intensities is insufficient, whereas the structured front end produces a spike-time representation that supports substantially higher accuracy. N-MNIST presents a different input regime: raw event frames are already informative, but the binned event-count latency representation provides the best performance among the tested configurations. Thus, the static and event-based front ends perform different input-specific transformations, while providing the backbone with a common TTFS representation.

The routing experiments then clarify how this representation should be propagated to the readout. The preserved representation \(P\) remains the main evidence source, whereas the intermediate and deep representations \(I\) and \(D\) are most useful through the sparse contributions defined by the residual and agreement mechanisms. In particular, preserving \(P\) is more effective than replacing it as processing depth increases, while \(\Delta_{\mathrm{res}}\) and \(\Delta_{\mathrm{agree}}\) provide additional evidence without discarding the early code. The decision-level analysis further shows that these contributions modify only a subset of the decisions made from \(P\), repairing part of the remaining errors rather than broadly changing the classifier output.

\section{Concluding Remarks}
\label{sec:conclusions}

This work proposes an original SNN multi-layer decision model amenable to online and localized STDP/R-STDP based training. It encodes image and visual streaming data into latency representations that are processed via multiple convolutional SNN layers, trained in an online fashion. A final readout layer, trained with a reward-modulated STDP mechanism, performs the final classification task. 

Our primary architectural innovation combines concepts from residual connections,  inception and consensus into a multi-layer SNN backbone. This design preserves early latency codes while progressively refining them, ensuring critical spike-time evidence remains available across deeper layers. 
Experiments show that preserving and refining the early temporal code is preferable to progressively replacing it with deeper representations. Across the considered datasets, the early code remains the main source of discriminative evidence, while the additional refinements become increasingly useful on more complex data. A decision-level analysis shows that these contributions affect only a subset of predictions, {\it repairing} part of the residual errors rather than broadly altering the decisions produced from the early code. Moreover, an activity-budget analysis shows that much of the classification performance is retained after removing a substantial fraction of the later or weaker events presented to the readout.
On the higher-variability visual tasks, the performance of the so obtained SNN consistently surpass existing STDP/R-STDP baselines, positioning our framework as a viable and effective approach for online learning scenarios.

Future work can investigate whether the same temporal-routing principle extends to larger event-based tasks, recurrent architectures, and neuromorphic hardware, performing direct activity and energy measurements.

\section*{CRediT authorship contribution statement}
\textbf{Aidin Attar}: Conceptualization, Methodology, Software, Investigation, Writing -- original draft, Writing -- review \& editing.\newline
\textbf{Eleonora Cicciarella}: Methodology, Writing -- review \& editing.\newline
\textbf{Michele Rossi}: Supervision, Writing -- review \& editing.

\section*{Declaration of competing interest}

The authors declare that they have no known competing financial interests or personal relationships that could have appeared to influence the work reported in this paper.

\section*{Declaration of generative AI usage}
During the preparation of this work, the authors used OpenAI's ChatGPT to assist with language refinement and readability. After using this tool, the authors reviewed and edited the content as needed and take full responsibility for the content of the publication.

\bibliographystyle{model1-num-names}
\bibliography{references}

\clearpage
\appendix

\section{Reproducibility Details}
\label{app:reproducibility}

Tables~\ref{tab:shared_parameters}-\ref{tab:readout_parameters} report the numerical settings of the main four-layer models. Static images and event streams use different deterministic encoders, after which the same layer-wise backbone, sparse MDTF module, and temporal R-STDP readout are applied. In the following tables, $k/s/p$ denotes kernel size, stride, and padding;
Top-$k_{\mathrm{map}}$ and Top-$k_{\mathrm{fus}}$ denote the map-wise and
fusion sparsity caps, respectively; Ep. denotes the number of training epochs.

\begin{table}[!ht]
    \centering
    \scriptsize
    \renewcommand{\arraystretch}{1.05}
    \setlength{\tabcolsep}{4pt}
    \caption{Front-end and shared optimization settings. The whitening stride is used only to sample fitting patches; the learned whitening operator is applied densely with unit stride. Static calibration is fitted on the training split and held fixed thereafter.}
    \label{tab:shared_parameters}
    \begin{tabular}{p{0.12\textwidth}p{0.18\textwidth}p{0.21\textwidth}p{0.21\textwidth}p{0.18\textwidth}}
        \toprule
        \textbf{Component} & \textbf{Parameter} & \textbf{MNIST/Fashion-MNIST} & \textbf{CIFAR-10} & \textbf{N-MNIST} \\
        \midrule
        Input source & Raw data & grayscale images & RGB images & polarity event stream \\
        Latency encoder & Representation & calibrated local-decorrelation polarity TTFS & calibrated local-decorrelation polarity TTFS & event-map TTFS \\
        Latency encoder & Output channels & 2 & 6 & 10 \\
        Local decorrelation & fit-patch $k$, stride, $\epsilon$ & $7$, $2$, $10^{-2}$ & $9$, $2$, $10^{-2}$ & -- \\
        Local decorrelation & PCA component fraction, max. patches, fit batch & $1.0$, $10^6$, $64$ & $1.0$, $10^6$, $64$ & -- \\
        Polarity calibration & split and scaling & positive/negative; training-set min--max per channel and location & positive/negative; training-set min--max per channel and location & native ON/OFF channels \\
        Signed-context gate & patch, $\alpha$, loser factor, mixed-ratio interval & $5$, $.15$, $.25$, $[.2,.7]$ & disabled & -- \\
        Polarity reweighting & patch, dominant compression/weak boost, margin & $4$, $.10/.15$, $.10$ & disabled & -- \\
        Pair compensation & patch, $\alpha$, loser factor, negative bias, margin & $6$, $.10$, $.25$, $.15$, $.10$; energy gate $[.6,.9]$ & $6$, $.10$, $.25$, $.15$, $.10$; energy gate disabled & -- \\
        Event map & temporal bins and denoising & -- & -- & $5\times2$ polarities; $10\,000\,\mu\mathrm{s}$ \\
        Event map & response normalization & -- & -- & log1p; local radius $2$, $\epsilon=10^{-4}$; sample-maximum normalization; $\gamma=1$ \\
        Convolutional STDP & weight bounds and target time & $[0,1]$, $.95$ & $[0,1]$, $.95$ & $[0,1]$, $.95$ \\
        $S_1$ threshold schedule & initial rate, minimum, annealing & $1$, $2$, $.95$ & $1$, $2$, $.95$ & $0$, $.2$, $.95$ \\
        $S_2$ threshold schedule & initial rate, minimum, annealing & $1$, $4$, $.95$ & $1$, $4$, $.95$ & $1$, $4$, $.95$ \\
        $S_{3:4}$ threshold schedule & initial rate, minimum, annealing & $0$, $0$, $1$ & $0$, $0$, $1$ & $0$, $0$, $1$ \\
        Weight initialization & $S_1$; $S_2$ & normal $.5\pm.01$; normal $.3\pm.01$ & normal $.5\pm.01$; normal $.3\pm.01$ & normal $.5\pm.01$; normal $.3\pm.01$ \\
        Weight initialization & $S_3$; $S_4$ & identity-centered, $g=.45$, $\sigma=.01$ & identity-centered, $g=.45$, $\sigma=.01$ & identity-centered, $g=.45$, $\sigma=.01$ \\
        Readout initialization & mean, std., bounds; normalization & $.3$, $.01$, $[0,1]$; fixed $\ell_1$ norm & $.3$, $.01$, $[0,1]$; fixed $\ell_1$ norm & $.3$, $.01$, $[0,1]$; fixed $\ell_1$ norm \\
        Readout regularizer & threshold rate and annealing & $5$, $.5$ & $5$, $.5$ & $5$, $.5$ \\
        Readout stability & minimum active output neurons & 36 & 36 & 36 \\
        \bottomrule
    \end{tabular}
\end{table}

\begin{table}[!ht]
    \centering
    \tiny
    \renewcommand{\arraystretch}{1.05}
    \setlength{\tabcolsep}{2pt}
    \caption{Backbone and fusion parameters for the four-layer locally trained spiking pipeline. The output column reports the latency representation consumed by the following stage. The $C_2$ representations were exported with pooling kernel/stride $2/1$. The deep rows describe the adaptive path used by the final model.}
    \label{tab:architecture_parameters}
    \resizebox{\textwidth}{!}{
    \begin{tabular}{llllllllll}
        \toprule
        \textbf{Block} &
        \textbf{Setting} &
        \textbf{Input} &
        \textbf{Output} &
        \textbf{$k/s/p$ or op} &
        \textbf{$\theta_0$ or $m_{\mathrm{agree}}$} &
        \textbf{Top-$k_{\mathrm{map/fus}}$} &
        \textbf{$g,\sigma$} &
        \textbf{$A^+/A^-/\beta$} &
        \textbf{Ep.} \\
        \midrule
        $S_1{+}C_1$ & MNIST/Fashion-MNIST & $2{\times}28{\times}28$ & $128{\times}6{\times}6$ & $5/1/0$; pool $4/4$ & 5.0 & 0 & - & $.100/.100/1.00$ & 8 \\
        $S_1{+}C_1$ & CIFAR-10 & $6{\times}32{\times}32$ & $128{\times}7{\times}7$ & $5/1/0$; pool $4/4$ & 10.0 & 0 & - & $.100/.100/1.00$ & 18 \\
        $S_1{+}C_1$ & N-MNIST & $10{\times}34{\times}34$ & $128{\times}7{\times}7$ & $5/1/0$; pool $4/4$ & 4.0 & 0 & - & $.100/.080/.85$ & 2 \\
        \addlinespace
        $S_2{+}C_2$ & MNIST/Fashion-MNIST & $128{\times}6{\times}6$ & $256{\times}5{\times}5$ & $1/1/0$; pool $2/1$ & 24.0 & 32 & - & $.100/.100/1.00$ & 3 \\
        $S_2{+}C_2$ & CIFAR-10 & $128{\times}7{\times}7$ & $256{\times}6{\times}6$ & $1/1/0$; pool $2/1$ & 24.0 & 32 & - & $.100/.100/1.00$ & 3 \\
        $S_2{+}C_2$ & N-MNIST & $128{\times}7{\times}7$ & $256{\times}6{\times}6$ & $1/1/0$; pool $2/1$ & 8.0 & 32 & - & $.100/.100/1.00$ & 2 \\
        \addlinespace
        $S_3$ & adaptive deep path & $256{\times}H{\times}W$ & $256{\times}H{\times}W$ & $1/1/0$ & 0.5 & 64 & $.45,.010$ & $.003/.003/.85$ & 2 \\
        $S_4$ & adaptive deep path & $256{\times}H{\times}W$ & $256{\times}H{\times}W$ & $1/1/0$ & 0.5 & 64 & $.45,.010$ & $.003/.003/.85$ & 2 \\
        \addlinespace
        Fusion & MNIST/Fashion-MNIST & $P:128{\times}6{\times}6$ + $2{\times}256{\times}5{\times}5$ & $d_{\mathrm{feat}}=17\,408$ & $[P,\Delta_{\mathrm{res}},\Delta_{\mathrm{agree}}]$ & $m_{\mathrm{agree}}=.001$ & $128/16$ & - & - & - \\
        Fusion & CIFAR-10 & $P:128{\times}7{\times}7$ + $2{\times}256{\times}6{\times}6$ & $d_{\mathrm{feat}}=24\,704$ & $[P,\Delta_{\mathrm{res}},\Delta_{\mathrm{agree}}]$ & $m_{\mathrm{agree}}=.001$ & $128/16$ & - & - & - \\
        Fusion & N-MNIST & $P:128{\times}7{\times}7$ + $2{\times}256{\times}6{\times}6$ & $d_{\mathrm{feat}}=24\,704$ & $[P,\Delta_{\mathrm{res}},\Delta_{\mathrm{agree}}]$ & $m_{\mathrm{agree}}=.001$ & $128/16$ & - & - & - \\
        \bottomrule
    \end{tabular}
    }
\end{table}
\begin{table}[!ht]
    \centering
    \tiny
    \renewcommand{\arraystretch}{1.05}
    \setlength{\tabcolsep}{2pt}
    \caption{Temporal R-STDP readout operating points for the main reported models. All rows use additive STDP, fixed per-neuron $\ell_1$ weight normalization, global winner-take-all decoding by output spike time, $\tau_{\max}=1$, $m_{\mathrm{ro}}=.005$, prototype decay $.5$, target scale $2$, floor $0$, target guard $1.25$, anti-guard $0$, and learning-rate annealing $.75$. ``Updated prototypes'' gives the number selected from the target population and from each selected hard-negative class, respectively.}
    \label{tab:readout_parameters}
    \resizebox{\textwidth}{!}{
    \begin{tabular}{lrrrrrrrrr}
        \toprule
        \textbf{Dataset} &
        \textbf{$d_{\mathrm{feat}}$} &
        \textbf{$N_{\mathrm{out}}$} &
        \textbf{Proto./class} &
        \textbf{$\theta_0$} &
        \textbf{$K$} &
        \textbf{Updated proto.} &
        \textbf{Non-target scale} &
        \textbf{$\lambda_{\max}$} &
        \textbf{Correct scale}\\
        \midrule
        MNIST & $17\,408$ & 80 & 8 & 120 & 7 & $1/3$ & 0.25 & 0.0035 & 0.0075 \\
        Fashion-MNIST & $17\,408$ & 40 & 4 & 223 & 5 & $1/2$ & 0.32 & 0.0050 & 0.1000 \\
        CIFAR-10 & $24\,704$ & 40 & 4 & 400 & 5 & $1/2$ & 0.32 & 0.0050 & 0.1000 \\
        N-MNIST & $24\,704$ & 60 & 6 & 243 & 9 & $1/3$ & 0.35 & 0.0030 & 0.0050 \\
        \bottomrule
    \end{tabular}
    }
\end{table}
\end{document}